\documentclass[runningheads]{llncs}
\usepackage{graphicx}
\usepackage{amssymb, amsmath, bm, latexsym,comment}
\usepackage{multirow}
\usepackage{xcolor}
\usepackage{subfigure}
\usepackage{indentfirst}
\usepackage{setspace}
\usepackage{verbatim}
\usepackage{array}
\usepackage{arydshln}
\usepackage[misc]{ifsym} 
\usepackage[colorlinks,linkcolor=blue]{hyperref}
\usepackage{xcolor}
\usepackage[normalem]{ulem} 

\providecommand{\Leireftb}[1]{Table~\ref{#1}}
\providecommand{\Leireffig}[1]{Fig.~\ref{#1}}

\providecommand{\citep}[1]{\cite{#1}}

\begin{document}
\title{Deep Computational Model for the Inference of Ventricular Activation Properties}
\titlerunning{PS-DCM for the Inference of Ventricular Activation Properties}

\author{ 
Lei Li \inst{1}${^{(\textrm{\Letter})}}$ \and
Julia Camps \inst{2} \and 
Abhirup Banerjee \inst{1,3} \and 
Marcel Beetz \inst{1} \and 
Blanca Rodriguez \inst{2} \and 
Vicente Grau \inst{1} 
} 

\authorrunning{L. Li et al.}


\institute{
Department of Engineering Science, University of Oxford, Oxford, UK \and
Department of Computer Science, University of Oxford, Oxford, UK \and
Division of Cardiovascular Medicine, Radcliffe Department of Medicine, University of Oxford, Oxford, UK \\
\email{lei.li@eng.ox.ac.uk}
}

\maketitle 
\begin{abstract}
Patient-specific cardiac computational models are essential for the efficient realization of precision medicine and in-silico clinical trials using digital twins. 
Cardiac digital twins can provide non-invasive characterizations of cardiac functions for individual patients, and therefore are promising for the patient-specific diagnosis and therapy stratification. 
However, current workflows for both the anatomical and functional twinning phases, referring to the inference of model anatomy and parameter from clinical data, are not sufficiently efficient, robust, and accurate. 
In this work, we propose a deep learning based patient-specific computational model, which can fuse both anatomical and electrophysiological information for the inference of ventricular activation properties, i.e., conduction velocities and root nodes.
The activation properties can provide a quantitative assessment of cardiac electrophysiological function for the guidance of interventional procedures.
We employ the Eikonal model to generate simulated electrocardiogram (ECG) with ground truth properties to train the inference model, where specific patient information has also been considered.
For evaluation, we test the model on the simulated data and obtain generally promising results with fast computational time.

\end{abstract}

\section{Introduction}

Cardiovascular diseases are one of the most common diseases globally, resulting in changes in cardiac anatomy, structure, and function \citep{journal/Lancet/kaptoge2019}.
Ventricular activation properties offer a valuable quantitative description of electrical activation and propagation, which is essential for identifying arrhythmias, localizing diseased tissue, and stratifying patients at risk \citep{journal/MedIA/camps2021,journal/JCP/grandits2020}.
For example, the location of the Purkinje endocardial root nodes (RNs), i.e., earliest activation sites, can provide important information for the selection of optimal implantation sites of pacing leads \citep{journal/TBME/pashaei2011}.
Conduction velocities (CVs) can describe the speed and direction of electrical propagation through the heart, and its alterations play an important role in the generation and maintenance of cardiac arrhythmias \citep{journal/Cell/han2021}. 

Electrocardiogram (ECG) can provide a substantial amount of information about the heart rhythm and reveal abnormalities related to the conduction system.
For instance, the QRS morphology in 12-lead ECG can indicate the origin of ventricular activation, which can be used to guide the clinicians to the potential ablation targets in real time \citep{journal/PCE/park2012}.
However, there exist large inter-subject anatomical variations that may modify the ECG patterns for specific activation properties.
Besides, ECG can not be used to locate and characterize diseases, such as arrhythmias.
The cardiac structural and functional information from imaging data, such as ultrasound, computed tomography or cardiac magnetic resonance (CMR), could be complementary to the information provided by ECG.
Computational models combining ECG and imaging data can be used to estimate ventricular activation properties for therapy guidance and variability interpretation among different patients \citep{journal/IF/martinez2021,journal/NRC/niederer2019}.  
However, it is complicated to accurately localize RNs, as there is limited knowledge about the actual topology of Purkinje activation networks \citep{journal/ABE/gillette2021}.
Moreover, the localization is often computationally expensive due to the complexity of structural and spatial variations of such networks. 
The estimation of CVs is also fundamentally challenging owing to the underlying mechanisms of complex, nonlinear, and heterogeneous myocardial activation \citep{journal/TBME/good2021}.
The simultaneous inference of CVs and RNs could be more challenging considering the existence of continuous and discrete mixed-type parameter space.

Nevertheless, there exist several CV and RN estimation techniques in the literature \citep{journal/CBM/cantwell2015}.
For CV estimation, Bayly et~al. \citep{journal/TBME/bayly1998} utilized inverse-gradient techniques to predict CVs from epicardial mapping data for the understanding and description of reentrant arrhythmias.
Chinchapatnam et~al. \citep{journal/TMI/chinchapatnam2008} employed an adaptive algorithm for the estimation of local CVs from a noncontact mapping of the endocardial surface potential.
Instead of predicting endocardial CV, Good et~al. \citep{journal/TBME/good2021} considered both epicardial and volumetric CVs and examined triangulation-based, inverse-gradient-based, and streamline-based techniques for comparison.
Compared to CV estimation, the localization of RNs received limited attention so far with only a few works, some of which solely used ECG signals for the localization \citep{journal/JNMBE/grandits2021}. 
The simultaneous optimization of CVs and RNs has been explored, but it is generally achieved via conventional iterative algorithms \citep{journal/MedIA/camps2021,journal/TBE/giffard2018,journal/JCP/grandits2020,journal/EP/pezzuto2021}.
Recently, deep learning based methods have achieved promising performance for cardiac activation modeling \citep{journal/EP/bacoyannis2021,conf/STACOM/meister2020}.
For example, Bacoyannis et~al. \citep{journal/EP/bacoyannis2021} proposed a $\beta$-conditional variational autoencoder to predict activation maps for various cardiac geometries with corresponding simulated body surface potentials.
Meister et~al. \citep{conf/STACOM/meister2020} utilized a graph convolutional regression network to predict the activation time maps from ECG and CMR images.

In this work, we have proposed a patient-specific deep computational model (PS-DCM) for an efficient and simultaneous estimation of ventricular activation properties, i.e., RNs and CVs.
As the activation properties are unavailable in healthy subjects, we create ``virtual cohorts" by simulation via Eikonal models for known ground truth values.
We consider additional physiological information as conditions to model the variations among specific sub-populations.
Also, we analyze the relationship between sampling latent space and the predictions, i.e., anatomy, signal, and activation parameters.
To the best of our knowledge, this is the first deep learning based computational model for ventricular activation property estimation.

\section{Methodology}

\begin{figure*}[t]\center
 \includegraphics[width=0.88\textwidth]{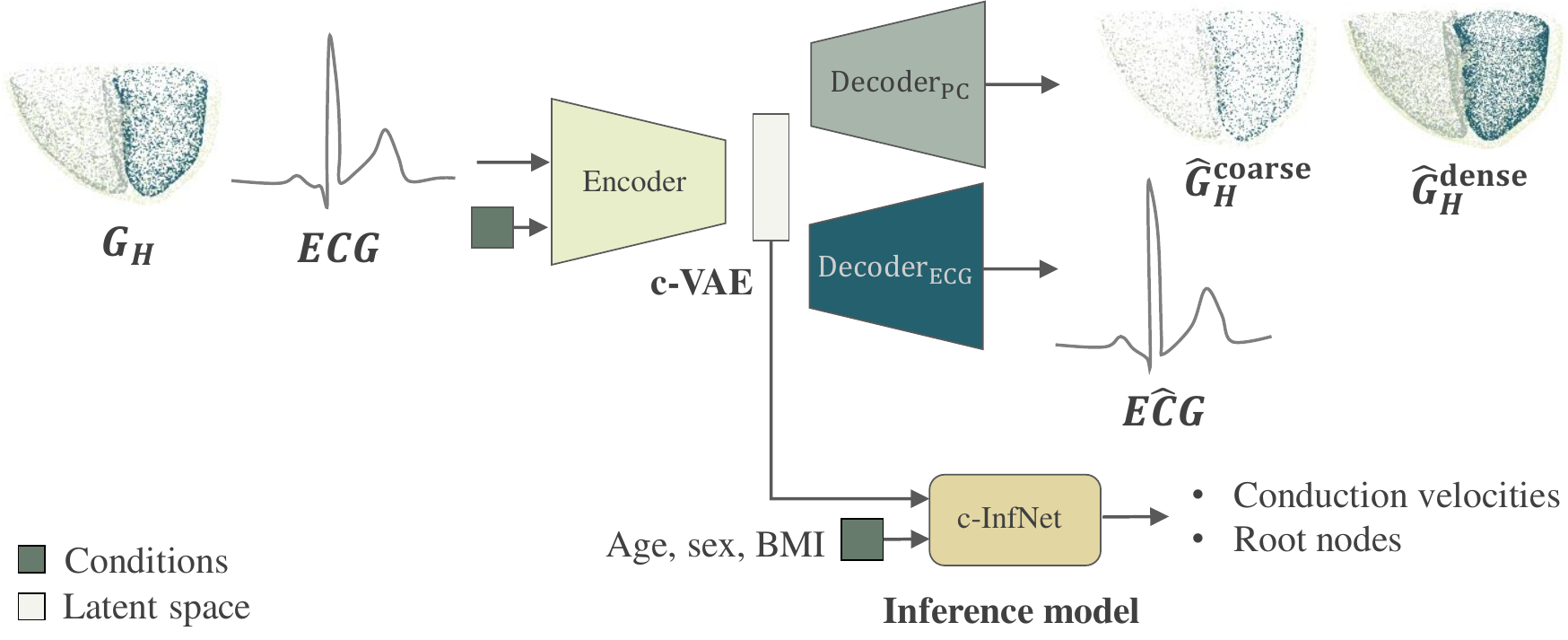}\\[-2ex]
   \caption{The proposed patient-specific deep computational model (PS-DCM) for the inference of ventricular activation properties. 
   The conditional variational autoencoder (c-VAE) aims to learn anatomical and electrical information that is used to assist the conditional inference network (c-InfNet) in predicting the activation parameters.}
\label{fig:method:framework}
\end{figure*}

Figure~\ref{fig:method:framework} provides an overview of the proposed PS-DCM, consisting of a conditional variational autoencoder (c-VAE) and an inference model.
Here, the c-VAE includes one encoder and two separated decoders for the structural and signal reconstructions, respectively, while
the conditional inference network (c-InfNet) aims to predict the ventricular activation properties based on the low-dimensional features from c-VAE.
In Sec.~\ref{method:mesh generation}, the mesh generation from 2D end-diastolic CMR images is introduced.
The generation of simulated data as the ground truth of the CVs and RNs is described in Sec.~\ref{method:simulation}.
Finally, Sec.~\ref{method:computational model} presents the details of the computational model for the prediction of two activation properties.

\subsection{Geometrical Triangular Mesh Generation} \label{method:mesh generation}

To obtain patient-specific 3D anatomical information, we generate smooth 3D biventricular tetrahedral meshes from 2D multi-view CMR images via a completely end-to-end automatic pipeline \citep{journal/PTRSA/banerjee2021}.
The pipeline starts by performing a deep learning based ventricle segmentation on the automatically selected slices from long- and short-axis CMR images.
Considering the potential misalignments due to respiration and cardiac motions, we employ the intensity and contour information as well as statistical shape model for in-plane and out-plane misalignment corrections, respectively.
Furthermore, we perform a surface mesh reconstruction from cardiac contours, which also mitigates the remaining discrepancies between sparse 3D contours.
For the following simulation of electrical activity in the heart, we further generate the 3D volumetric tetrahedral mesh from the 3D biventricular surface mesh using mesh generators \citep{journal/PE/engwirda2016,journal/ACM/si2015}.


\subsection{Simulated Data Generation via Eikonal Models} \label{method:simulation}

As the ventricular activation properties can not be measured for clinical data, we built a synthetic dataset from real clinical data via Eikonal model based simulation \citep{journal/MedIA/camps2021}.
Specifically, the Eikonal model is simulated over the generated 3D tetrahedral meshes in Sec.~\ref{method:mesh generation} and can be defined as,
\begin{equation}
	\sqrt{(\nabla d^{T}\cdot \nabla d)} = 1.
\end{equation}
Here, $\nabla d = v \nabla t$ where $\nabla t$ is the traveling time passing through a node and $v$ are the CVs of the fiber, sheet (transmural) and sheet-normal directions.
For fast optimization, we regard the tetrahedral mesh as a graph, where electric current can enter directly from two connected nodes through the edges connecting them.
Consequently, one could use Dijkstra's algorithm to predict Eikonal's activation time.
Note that we consider the position of the RN as the starting point with $t=0$, while the other nodes of the biventricular network are the destinations for which we need to compute $t$ for final activation time maps (ATMs).
Subsequently, one could calculate ECG signals from the ATMs simulated from Eikonal models via the pseudo-ECG equation with the electrode locations from torso geometry and orientation.
The pseudo-ECG equation is defined as follows,
\begin{equation}
\Phi_{e}=\sum_{j=1}^{N}-\left(\nabla V_{m}\right)_{j} \cdot\left[\nabla \frac{S_{j}}{r_{j}}\right],
\end{equation}
where $j$ is the index of tetrahedral element, $N$ is the number of tetrahedral elements, $\nabla V_{m}$ is an estimated gradient, $s$ is the normalized volume, and $r$ is distance calculated using the centroid of each element.
Here, we estimate the gradients by assigning $V_m^i=1$ if the node $i$ is activated, otherwise $V_m^i=0$, to generate a scaled amplitude ECG signal.
The pseudo-ECG method can efficiently generate normalized ECG signals without significant loss of morphological information compared to the bidomain simulation \citep{journal/FiP/minchole2019}.
As a result, with the Eikonal models one could generate virtual subjects by setting different CVs and RNs.

\subsection{Patient Specific Deep Computational Model} \label{method:computational model}

\begin{figure*}[t]\center
 \includegraphics[width=0.92\textwidth]{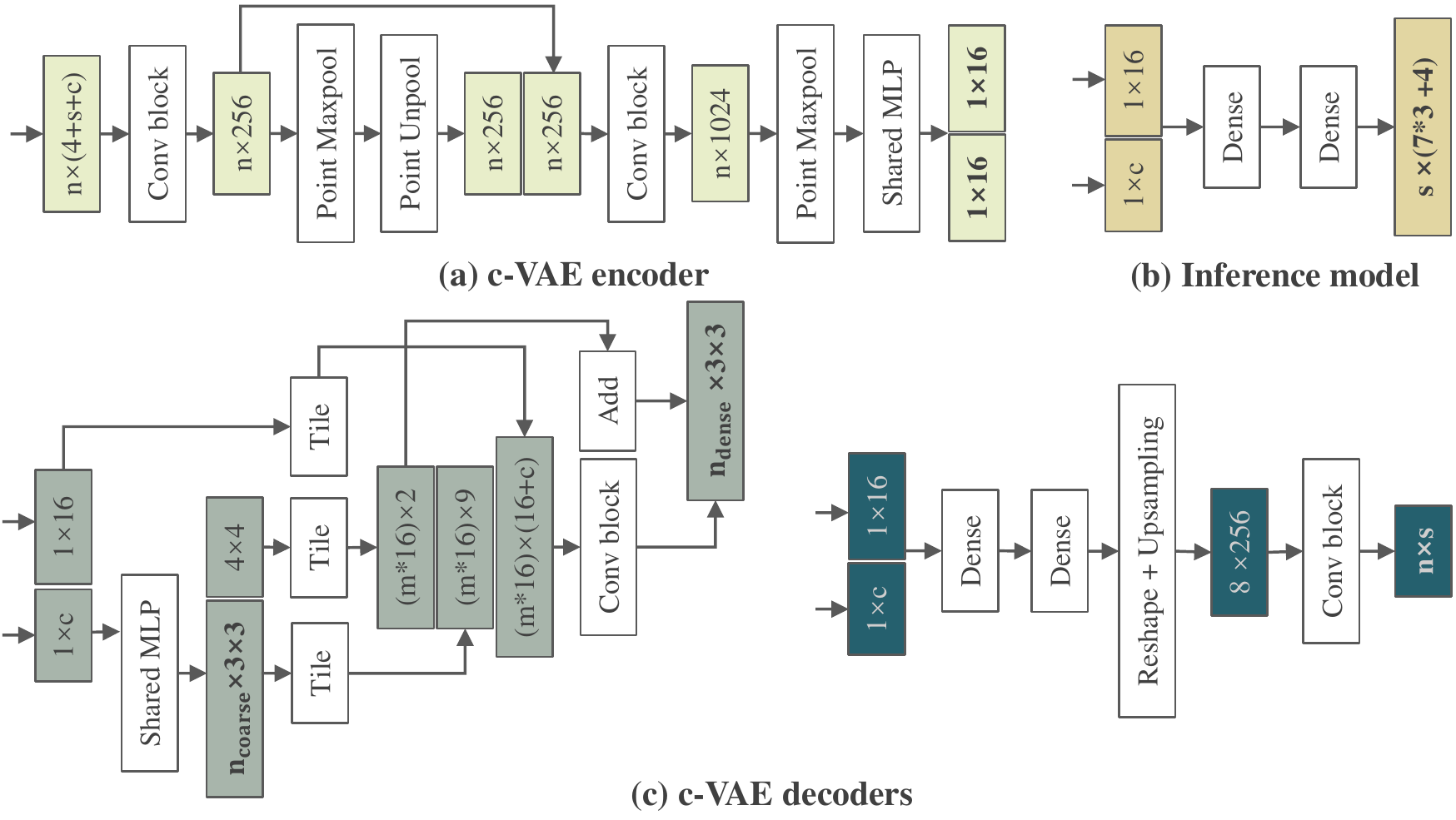}\\[-2ex]
   \caption{The network architecture of the proposed PS-DCM, where the output of each module is labeled using bold text. 
   Here, $n$ is the number of nodes of input point cloud (PC) which is a 4D vector (three point coordinates and a class label);
   $n_\text{coarse}$ and $n_\text{dense}$ are the numbers of nodes in the coarse and dense output PCs, respectively;
   $s$ is the number of simulated ECGs for each mesh ($n$ is set as the same value as the size of ECG signal for the convenience of input PC and ECG concatenation);
   $c$ is the number of conditions that has been reshaped to match the size of input PC for concatenation.
   }
\label{fig:method:network}
\end{figure*}

Figure~\ref{fig:method:network} presents the network architecture details of the proposed PS-DCM.
The c-VAE aims to reconstruct the anatomical and electrophysiological information, respectively.
The outputs of $\text{Decoder}_\text{PC}$ are coarse and dense point clouds (PCs) to simultaneously learn global shape and local structures of ventricles \citep{conf/STACOM/beetz2021}. The PC reconstruction loss function is defined as follows,
\begin{equation}
	\mathcal{L}^\text{rec}_\text{PC} = \sum_{i=1}^{K}\big(\mathcal{L}_{i,coarse}^\text{EMD} + \alpha \mathcal{L}_{i,dense}^\text{EMD}\big),
\end{equation}
where $K$ is the number of classes, $\alpha$ is the weight term between the two PCs, and $\mathcal{L}^{EMD}$ is the earth mover’s distance (EMD) \citep{conf/3DV/yuan2018} between the ground truth $G_H$ and predicted heart geometries $\hat{G}_H$,
\begin{equation}
	\mathcal{L}^{EMD}(G_H, \hat{G}_H) = \mathop{\arg\min}_{\xi: G_H \rightarrow \hat{G}_H} \sum_{p \in G_H}\|p-\xi(p)\|_{2},
\end{equation}
where $p$ is the node of point cloud and $\xi: G_H \rightarrow \hat{G}_H$ is a one-to-one correspondence.
$\text{Decoder}_\text{ECG}$ predicts the reconstructed ECG signals by minimizing the mean absolute error (MAE) between the ground truth and predicted ECG (denoted as $\text{E}\hat{\text{C}}\text{G}$),
\begin{equation}
	\mathcal{L}^\text{rec}_\text{ECG} = \mathcal{L}^\text{MAE}(\text{ECG}, \text{E}\hat{\text{C}}\text{G}).
\end{equation}
Therefore, the loss function for training the c-VAE could be calculated as,
\begin{equation}
	\mathcal{L}_\text{c-VAE} = \lambda_\text{PC}\mathcal{L}^\text{rec}_\text{PC} + \lambda_\text{ECG}\mathcal{L}^\text{rec}_\text{ECG} + \lambda_\text{KL}\mathcal{L}^\text{KL},
\end{equation}
where $\lambda_\text{PC}$, $\lambda_\text{ECG}$ and $\lambda_\text{KL}$ are balancing parameters, and $\mathcal{L}^\text{KL}$ is the Kullback-Leibler (KL) divergence loss to mitigate the distance between the prior and posterior distributions of the latent space.
Here, the posterior distribution is assumed as a standard normal distribution. 

As for the inference model, we predict the CVs and RNs based on the conditions and low-dimensional features learned from the c-VAE.
We also employ MAE and velocity-normalized (vnMAE) loss for the training of the inference model,
\begin{equation}
	\mathcal{L}_\text{inf} = \mathcal{L}^\text{MAE}_\text{RN} + \lambda_\text{CV}\mathcal{L}^\text{vnMAE}_\text{CV},
\end{equation}
where $\lambda_\text{CV}$ is a balancing parameter.
Hence, the total loss of the framework is defined by combining all the losses mentioned above,
\begin{equation}
	\mathcal{L}_\text{total} = \mathcal{L}_\text{c-VAE} + \lambda_\text{inf}\mathcal{L}_\text{inf},
\end{equation}
where $\lambda_\text{inf}$ is a balancing parameter.

\section{Experiment and Result}

\subsection{Materials}
\subsubsection{Data Acquisition and Pre-Processing}
We collect 100 subjects with cine CMR images and ECGs from the UK Biobank study \citep{journal/Nature/bycroft2018}.
Besides, we employ biological information, i.e., age, sex, and body mass index (BMI), as the conditions of cVAE, as they may affect the ECG morphology \citep{journal/OR/fraley2005,journal/PCE/taneja2001}.
For anatomical reconstruction, we resample all PCs into the PC with the same number of nodes, i.e., $n=2048$, and the numbers of nodes in the coarse and dense output PCs are set as 1024 and 4096.
For simulation, CVs are randomly selected within physiological ranges, i.e., fiber-directed, sheet-directed, sheet normal-directed, and endocardial-directed CVs are within the ranges [50, 88], [32, 49], [29, 45], and [120, 179] cm/s, respectively.
Moreover, we constrain the fiber-directed speed to be larger than the sheet-directed one, which in turn is larger than the sheet normal-directed speed \citep{journal/CAE/caldwell2009}.
In contrast, RNs are set to seven fixed homologous locations (3 in the right ventricle and 4 in the left ventricle) for realistic application \citep{journal/EP/cardone2016}.
Note that we did not consider ventricle-specific protocols for RN placement to ensure more flexible extension in the future.
For each mesh, we generate 10 virtual subjects with different ECGs but the same ventricle anatomy for the training of the PS-DCM.
We randomly split the data into three sets, i.e., 60 subjects for training, 10 subjects for validation and the remaining 30 subjects for the test.

\subsubsection{Gold Standard and Evaluation}
We employ specific ventricular activation properties to simulate data, and then real resampled PCs, simulated ECGs and corresponding properties are regarded as ground truth of this model.
We calculate the EMD from coarse and fine PCs with corresponding ground truth PCs, as the accuracy of PC reconstruction.
As there exist length variations among different leads of the simulated ECG signals, we performed a zero-pad to unify the length of ECG.
Therefore, in the evaluation of ECG reconstruction in this study, we only consider the non-zero signals when employing L1 loss as the discrepancy metric for the ECG data.
As for the inference model evaluation, we employ the average distance of root nodes and a velocity-normalized error metric.

\subsubsection{Implementation} 
The framework was implemented in PyTorch, running on a computer with 3.50 GHz Intel(R) Xeon(R) E-2146G CPU and an NVIDIA GeForce RTX 3060. 
We use the Adam optimizer to update the network parameters (weight decay = 1e-3). 
The initial learning rate is set to 1e-4 and multiplied by 0.7 every 30000 iterations. 
The balancing parameters in Sec.~\ref{method:computational model} are set as follows: $\alpha=0.1$, $\lambda_\text{PC}=0.1$, $\lambda_\text{ECG}=0.1$, $\lambda_\text{KL}=0.02$, $\lambda_\text{CV}=0.2$, and $\lambda_\text{inf}=0.01$.
The training of the model took about 2 hours (1000 epochs in total), while the inference of the networks required about 9 s to process one test image.

\subsection{Result}

\begin{table*} [t] \center
    \caption{
    Summary of the quantitative results of ventricular activation properties. LV: left ventricle; RV: right ventricle; C: conditions.
     }
\label{tb:exp:parameter result}
{\scriptsize	
\begin{tabular}{p{2.2cm}|p{1.4cm}p{1.4cm}|p{1.4cm}p{1.4cm}p{1.7cm}p{1.5cm}} 
\hline
\multirow{2}*{Method} & \multicolumn{2}{c|}{Root Node Error (cm)} & \multicolumn{4}{c}{Conduction Velocity Error (\%)}\\
\cline{2-7}
~ & LV & RV & Fiber & Sheet & Sheet-Normal & Endocardial \\
\hline
PS-DCM w/o C    & $ 3.83 \pm 1.10  $ & $ 3.90 \pm 1.07 $  & $ 18.3 \pm 4.47 $ & $ 22.2 \pm 6.19 $ & $ 22.3 \pm 7.27 $ & $ 12.1 \pm 1.91 $  \\
PS-DCM w/o PC   & $ 4.13 \pm 1.07  $ & $ 3.75 \pm 1.06 $  & $ 24.3 \pm 6.26 $ & $ 29.7 \pm 14.7 $ & $ 38.8 \pm 20.6 $ & $ 14.8 \pm 4.79 $  \\
\textbf{PS-DCM} & $ 2.63 \pm 0.909 $ & $ 2.56 \pm 0.882 $ & $ 13.9 \pm 2.70 $ & $ 9.53 \pm 3.13 $ & $ 15.1 \pm 2.86 $ & $ 11.7 \pm 2.23 $  \\
\hline
\end{tabular} }\\
\end{table*}

\subsubsection{Inference Accuracy of Ventricular Activation Properties} 
\begin{figure*}[!t]\center
	\includegraphics[width=0.84\textwidth]{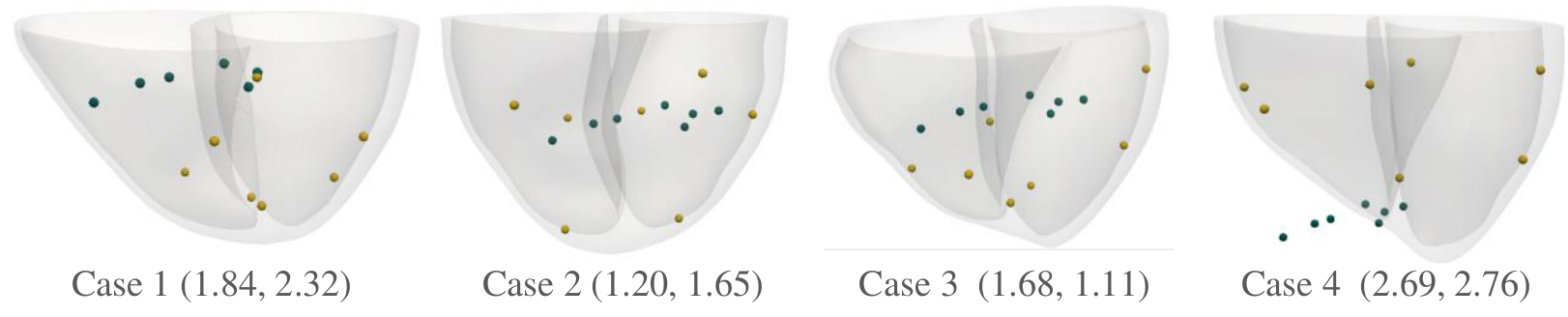}
	\caption{Visualization of ground truth (yellow spheres) and predicted root nodes (cyan-blue spheres) by the proposed method from three randomly selected cases. Here, we put the RN prediction errors on the LV and RV in brackets.}
	\label{fig:exp:RN}
\end{figure*}

\Leireftb{tb:exp:parameter result} presents the quantitative results of different methods for RN and CV inference.
One can see that the proposed PS-DCM method obtains the best inference results compared to the other two schemes without condition or PC reconstruction constraints, respectively.
It reveals the importance of anatomy and physical information for the patient-specific activation property prediction.
Even though we consider this information in our proposed framework, it is too challenging to predict the positions of RNs, as shown in \Leireffig{fig:exp:RN}.
One can see that the predicted RNs are generally located inside the middle (or even outside) of the ventricles instead of locally matching with the ground truth RNs.
It indicates that the anatomical constrains are desired for the deep learning based RN inference in the future. 
In contrast, the prediction of CVs is more promising with comparable results as the conventional method \citep{journal/MedIA/camps2021}, and there exist some accuracy variations for different directions of CVs.
Specifically, the sheet and endocardial-directed CVs are better identified than the CVs in fiber and sheet-normal directions by the proposed model.
For healthy subjects, the sheet and endocardial-directed CVs have been regarded as the dominant factors in the activation sequence patterns, while the impact from fiber and sheet-normal CVs is negligible \citep{journal/Cir/durrer1970}.
Consequently, there exist performance differences in CV inference from ECG data under healthy sinus rhythm conditions. 
The fiber-directed CV may play a more important role in pathological conditions, which, however, is out of the scope of this study.

\subsubsection{Reconstruction Quality of Point Cloud and ECG} 
\begin{figure*}[!t] \center
 \includegraphics[width=1\textwidth]{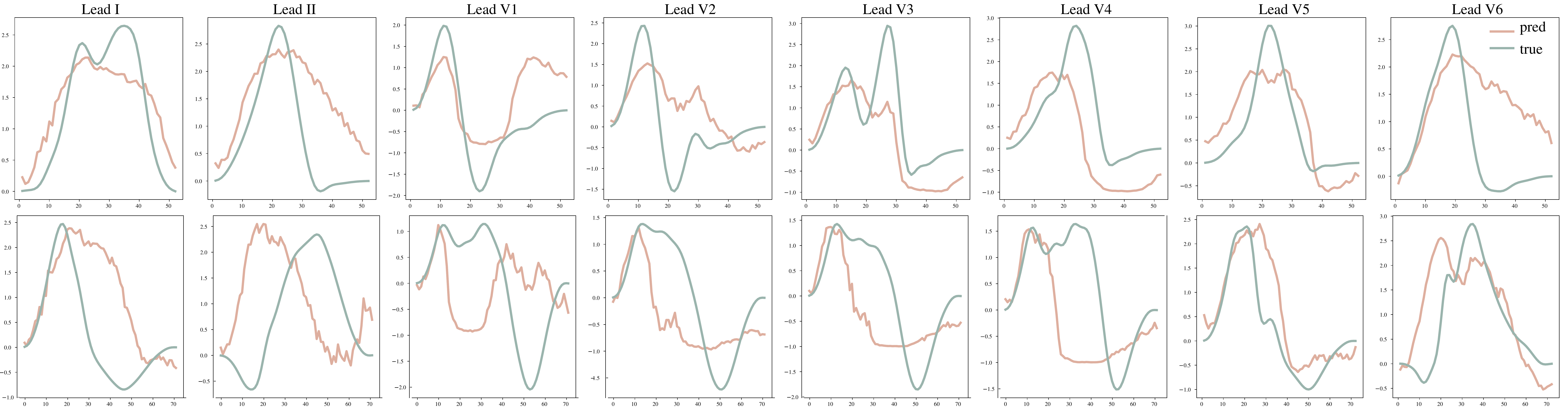}\\[-2ex]
   \caption{Illustration of reconstructed ECGs (labeled in pink) with corresponding simulated ground truth ECGs (labeled in cyan-blue).}
\label{fig:exp:ECG result}
\end{figure*}

The average PC reconstruction errors of the proposed method are $ 4.34 \pm 2.23$~cm and $ 4.07 \pm 2.08$~cm for coarse and fine PCs, respectively.
Figure~\ref{fig:exp:ECG result} presents the ECG visualization results from eight leads.
Note that in this study we only consider the eight independent leads instead of complete 12-lead ECG, as the remaining leads are linear combinations of the other leads \citep{journal/MedIA/camps2021}.
One can see that compared to the ground truth, the reconstructed ECGs are generally matched with input ECGs in several leads even though not quite smooth.
There exists reconstruction accuracy variance among different leads, namely the prediction of leads V1, V2, and V3 had larger misalignment compared to other leads.
We argue that simultaneous feature encoding of PCs and ECGs may be too challenging, resulting poor reconstruction results.
In the future, one may could consider employ separated encoders to extract features from PCs and ECGs \citep{journal/FiP/beetz2022}.

\subsubsection{Correlation Study}
\begin{figure*}[!t] \center
    \subfigure[] {\includegraphics[width=0.48\textwidth]{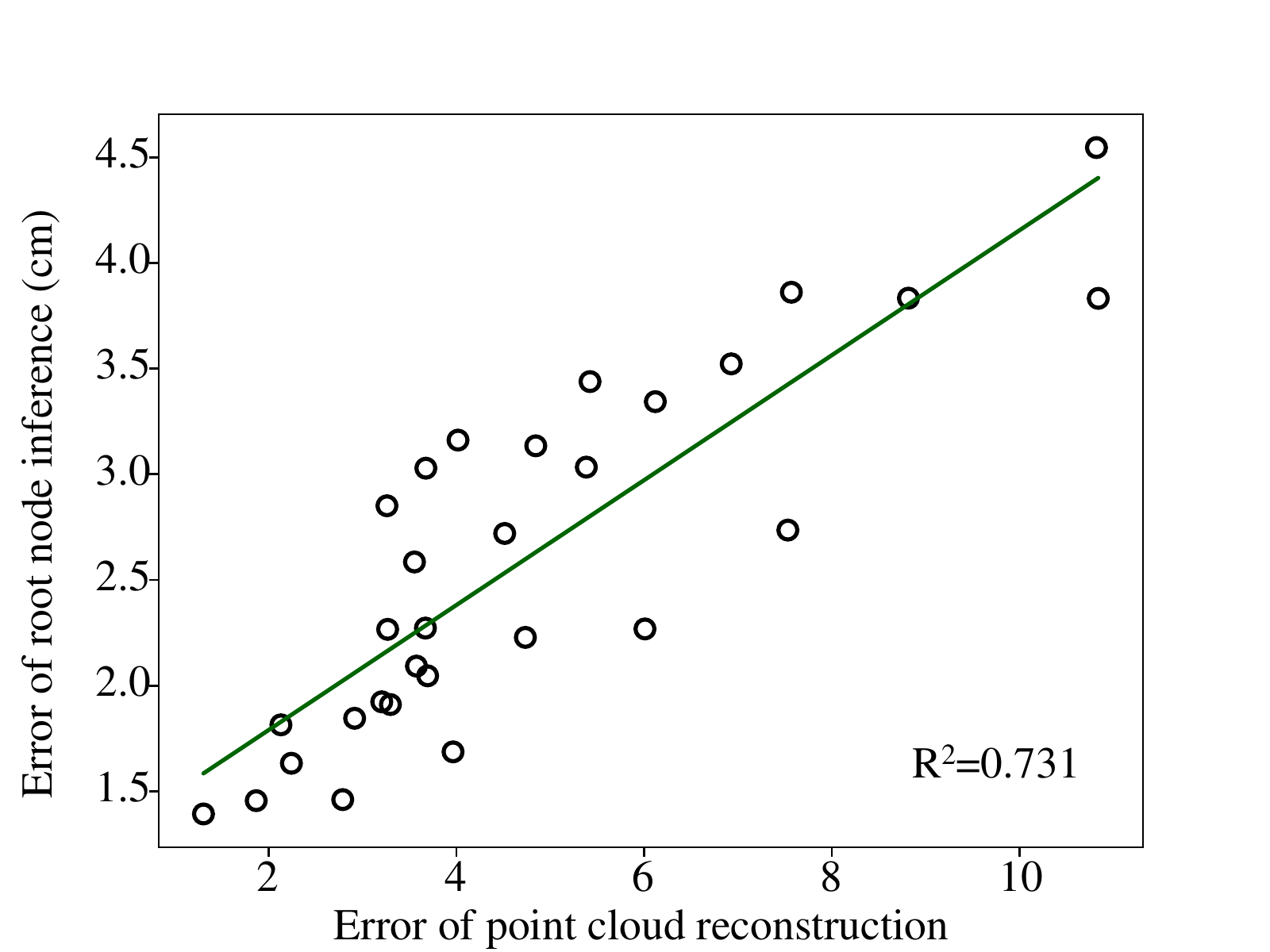}} 
    \subfigure[] {\includegraphics[width=0.48\textwidth]{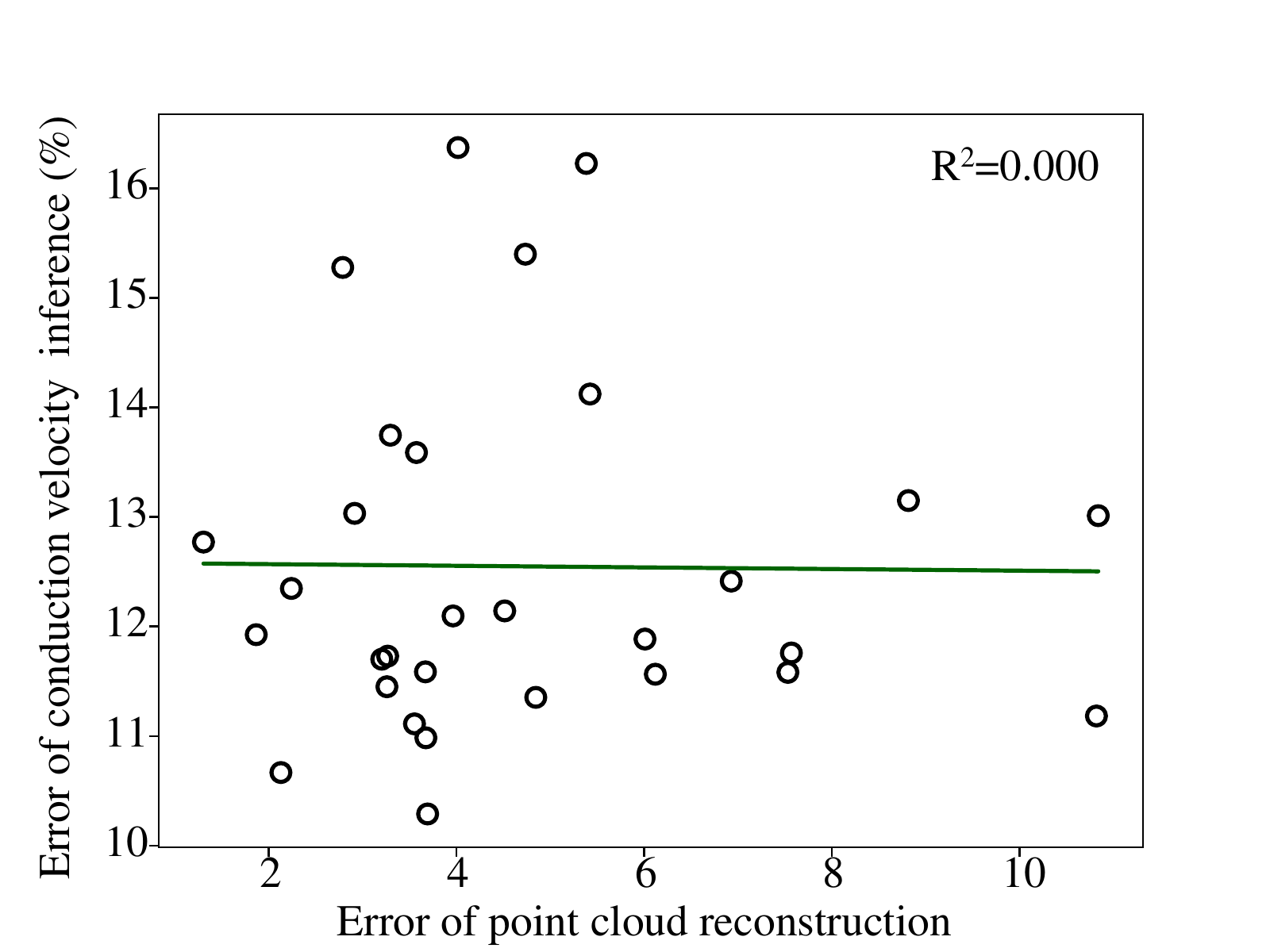}}
    \subfigure[] {\includegraphics[width=0.482\textwidth]{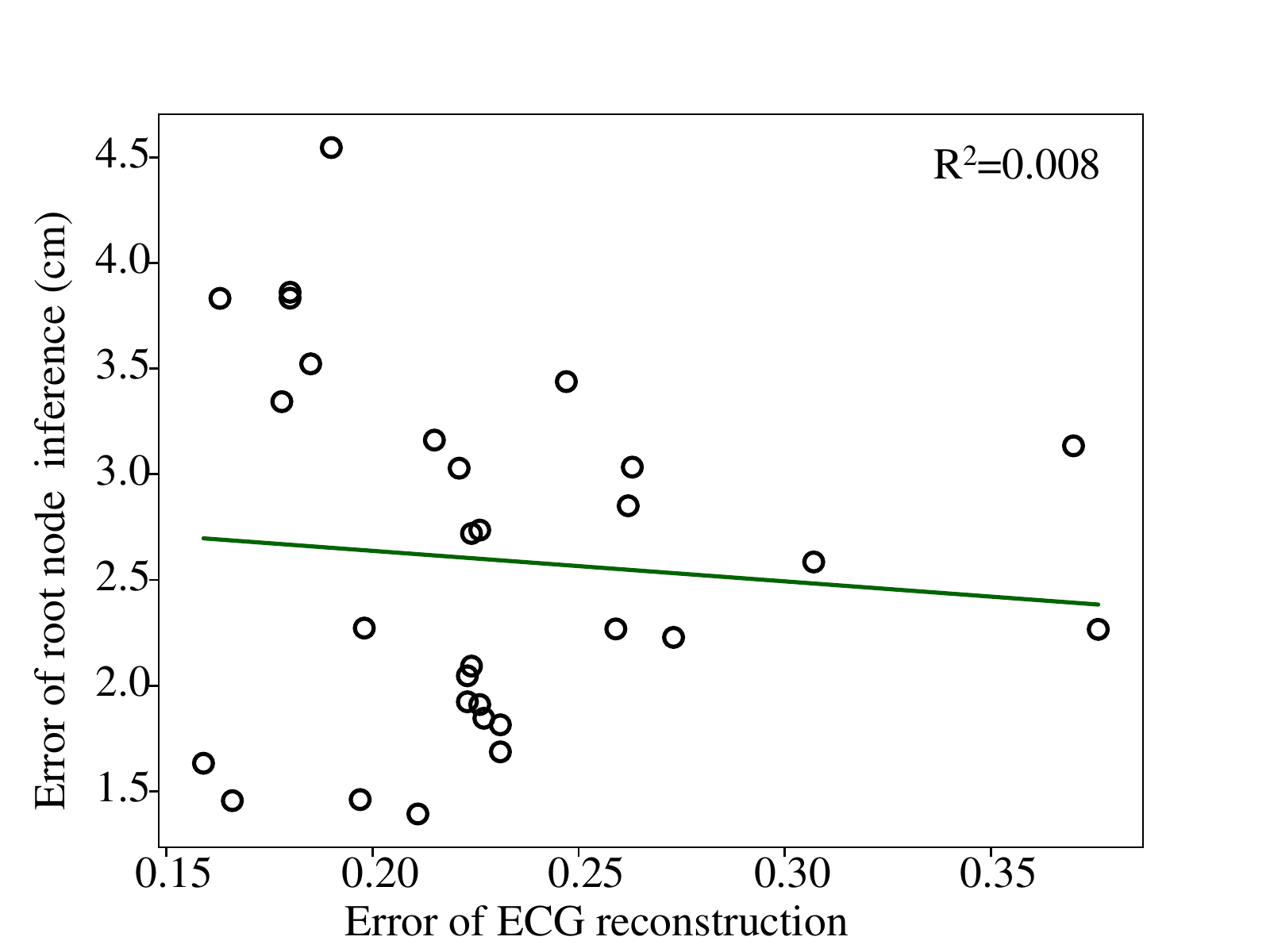}}
    \subfigure[] {\includegraphics[width=0.48\textwidth]{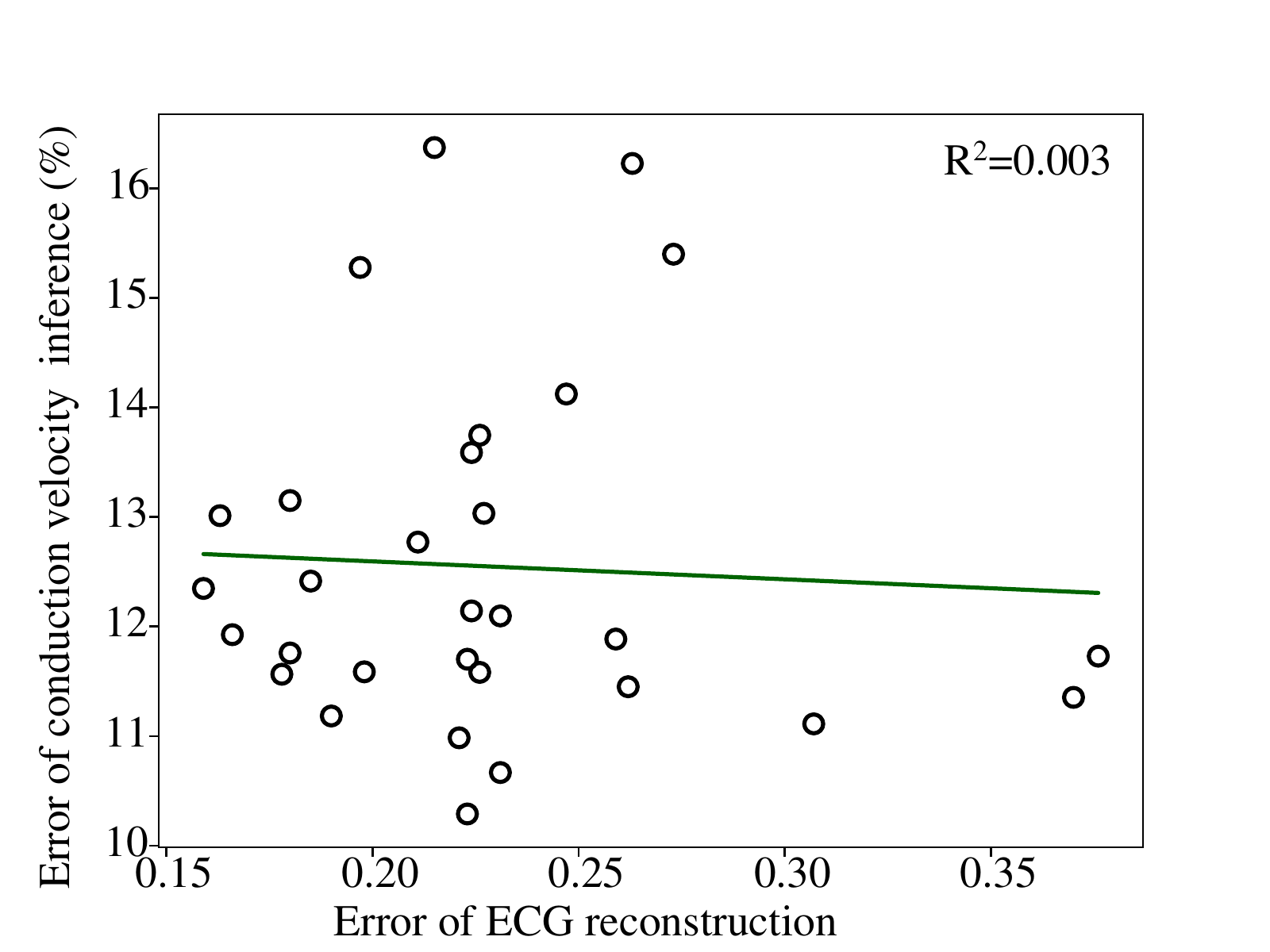}}
	\caption{
		Scatter plots along with the fitted linear regression lines,
		depicting the correlations between inference and reconstruction errors.
		}
	\label{fig:exp:scatter_plot}
\end{figure*}

To evaluate the robustness of the proposed inference scheme to the reconstruction error, we analyze the relationship between the reconstruction and inference errors by the proposed method.
We plot these two values for each test data as two-dimension scatter points along with the fitted linear regression, as presented in \Leireffig{fig:exp:scatter_plot}.
The $R^2$ values are estimated as 0.731, 0.000, 0.008, and 0.003 for PC-RN, PC-CV, ECG-RN, and ECG-CV correlations, respectively, indicating low linear correlations between inference and reconstruction accuracy except for PC-RN.
It implies that the CV inference by the proposed method does not rely on accurate PC/ECG reconstruction results.
In contrast, the RN inference may require accurate anatomy information from PC reconstruction but without the high demand for electrophysiological information.
This is reasonable, as RNs belong to ventricle positions that are locally distributed on the endocardium while CVs are more general properties. 

\section{Conclusion}
In this work, we have presented an end-to-end deep learning based computational model for simultaneous inference of RNs and CVs combining the ECG and CMR images.
The proposed algorithm has been applied to 100 simulated ECGs with the corresponding anatomies obtained from the UK biobank dataset.
The results have demonstrated the potential of an efficient activation property inference based on the physical information and low-dimensional features from the c-VAE.
Note that this is only a quite primary study with several limitations, such as assuming a known set of RNs and anisotropic CVs on specific directions (fiber, sheet, sheet-normal, and endocardial directions).
Moreover, currently we only consider cardiac anatomical information and ignore the torso geometry, which provides important information during the propagation of electrical conduction.
In the future, we will extend this work by including a more realistic representation of the cardiac conduction system.
Consequently, the developed models and techniques will enable further research in non-invasive personalization of the ventricular activation sequences for real patients.

\subsubsection{Acknowledgement.}
This work was funded by the CompBioMed 2 Centre of Excellence in Computational Biomedicine (European Commission Horizon 2020 research and innovation programme, grant agreement No. 823712).
L. Li was partially supported by the SJTU 2021 Outstanding Doctoral Graduate Development Scholarship.

\bibliographystyle{splncs04}
\bibliography{A_refs}

\begin{thebibliography}{10}
\providecommand{\url}[1]{\texttt{#1}}
\providecommand{\urlprefix}{URL }
\providecommand{\doi}[1]{https://doi.org/#1}

\bibitem{journal/EP/bacoyannis2021}
Bacoyannis, T., Ly, B., Cedilnik, N., Cochet, H., Sermesant, M.: Deep learning
  formulation of electrocardiographic imaging integrating image and signal
  information with data-driven regularization. EP Europace
  \textbf{23}(Supplement\_1),  i55--i62 (2021)

\bibitem{journal/PTRSA/banerjee2021}
Banerjee, A., Camps, J., Zacur, E., Andrews, C.M., Rudy, Y., Choudhury, R.P.,
  Rodriguez, B., Grau, V.: A completely automated pipeline for {3D}
  reconstruction of human heart from {2D} cine magnetic resonance slices.
  Philosophical Transactions of the Royal Society A  \textbf{379}(2212),
  20200257 (2021)

\bibitem{journal/TBME/bayly1998}
Bayly, P.V., KenKnight, B.H., Rogers, J.M., Hillsley, R.E., Ideker, R.E.,
  Smith, W.M.: Estimation of conduction velocity vector fields from epicardial
  mapping data. IEEE Transactions on Biomedical Engineering  \textbf{45}(5),
  563--571 (1998)

\bibitem{conf/STACOM/beetz2021}
Beetz, M., Banerjee, A., Grau, V.: Generating subpopulation-specific
  biventricular anatomy models using conditional point cloud variational
  autoencoders. In: International Workshop on Statistical Atlases and
  Computational Models of the Heart. pp. 75--83. Springer (2021)

\bibitem{journal/FiP/beetz2022}
Beetz, M., Banerjee, A., Grau, V.: Multi-domain variational autoencoders for
  combined modeling of mri-based biventricular anatomy and {ECG}-based cardiac
  electrophysiology. Frontiers in physiology p.~991 (2022)

\bibitem{journal/Nature/bycroft2018}
Bycroft, C., Freeman, C., Petkova, D., Band, G., Elliott, L.T., Sharp, K.,
  Motyer, A., Vukcevic, D., Delaneau, O., O’Connell, J., et~al.: The {UK
  B}iobank resource with deep phenotyping and genomic data. Nature
  \textbf{562}(7726),  203--209 (2018)

\bibitem{journal/CAE/caldwell2009}
Caldwell, B.J., Trew, M.L., Sands, G.B., Hooks, D.A., LeGrice, I.J., Smaill,
  B.H.: Three distinct directions of intramural activation reveal nonuniform
  side-to-side electrical coupling of ventricular myocytes. Circulation:
  Arrhythmia and Electrophysiology  \textbf{2}(4),  433--440 (2009)

\bibitem{journal/MedIA/camps2021}
Camps, J., Lawson, B., Drovandi, C., Minchole, A., Wang, Z.J., Grau, V.,
  Burrage, K., Rodriguez, B.: Inference of ventricular activation properties
  from non-invasive electrocardiography. Medical Image Analysis  \textbf{73},
  102143 (2021)

\bibitem{journal/CBM/cantwell2015}
Cantwell, C.D., Roney, C.H., Ng, F.S., Siggers, J.H., Sherwin, S.J., Peters,
  N.S.: Techniques for automated local activation time annotation and
  conduction velocity estimation in cardiac mapping. Computers in Biology and
  Medicine  \textbf{65},  229--242 (2015)

\bibitem{journal/EP/cardone2016}
Cardone-Noott, L., Bueno-Orovio, A., Minchol{\'e}, A., Zemzemi, N., Rodriguez,
  B.: Human ventricular activation sequence and the simulation of the
  electrocardiographic {QRS} complex and its variability in healthy and
  intraventricular block conditions. EP Europace  \textbf{18}(suppl\_4),
  iv4--iv15 (2016)

\bibitem{journal/TMI/chinchapatnam2008}
Chinchapatnam, P., Rhode, K.S., Ginks, M., Rinaldi, C.A., Lambiase, P., Razavi,
  R., Arridge, S., Sermesant, M.: Model-based imaging of cardiac apparent
  conductivity and local conduction velocity for diagnosis and planning of
  therapy. IEEE Transactions on Medical Imaging  \textbf{27}(11),  1631--1642
  (2008)

\bibitem{journal/Cir/durrer1970}
Durrer, D., Van~Dam, R.T., Freud, G., Janse, M., Meijler, F., Arzbaecher, R.:
  Total excitation of the isolated human heart. Circulation  \textbf{41}(6),
  899--912 (1970)

\bibitem{journal/PE/engwirda2016}
Engwirda, D.: Conforming restricted {D}elaunay mesh generation for piecewise
  smooth complexes. Procedia Engineering  \textbf{163},  84--96 (2016)

\bibitem{journal/OR/fraley2005}
Fraley, M., Birchem, J., Senkottaiyan, N., Alpert, M.: Obesity and the
  electrocardiogram. Obesity Reviews  \textbf{6}(4),  275--281 (2005)

\bibitem{journal/TBE/giffard2018}
Giffard-Roisin, S., Delingette, H., Jackson, T., Webb, J., Fovargue, L., Lee,
  J., Rinaldi, C.A., Razavi, R., Ayache, N., Sermesant, M.: Transfer learning
  from simulations on a reference anatomy for {ECGI} in personalized cardiac
  resynchronization therapy. IEEE Transactions on Biomedical Engineering
  \textbf{66}(2),  343--353 (2018)

\bibitem{journal/ABE/gillette2021}
Gillette, K., Gsell, M.A., Bouyssier, J., Prassl, A.J., Neic, A., Vigmond,
  E.J., Plank, G.: Automated framework for the inclusion of a his--purkinje
  system in cardiac digital twins of ventricular electrophysiology. Annals of
  Biomedical Engineering  \textbf{49}(12),  3143--3153 (2021)

\bibitem{journal/TBME/good2021}
Good, W.W., Gillette, K.K., Zenger, B., Bergquist, J.A., Rupp, L.C., Tate, J.,
  Anderson, D., Gsell, M.A., Plank, G., MacLeod, R.S.: Estimation and
  validation of cardiac conduction velocity and wavefront reconstruction using
  epicardial and volumetric data. IEEE Transactions on Biomedical Engineering
  \textbf{68}(11),  3290--3300 (2021)

\bibitem{journal/JNMBE/grandits2021}
Grandits, T., Effland, A., Pock, T., Krause, R., Plank, G., Pezzuto, S.:
  {GEASI}: Geodesic-based earliest activation sites identification in cardiac
  models. International Journal for Numerical Methods in Biomedical Engineering
   \textbf{37}(8),  e3505 (2021)

\bibitem{journal/JCP/grandits2020}
Grandits, T., Gillette, K., Neic, A., Bayer, J., Vigmond, E., Pock, T., Plank,
  G.: An inverse eikonal method for identifying ventricular activation
  sequences from epicardial activation maps. Journal of Computational Physics
  \textbf{419},  109700 (2020)

\bibitem{journal/Cell/han2021}
Han, B., Trew, M.L., Zgierski-Johnston, C.M.: Cardiac conduction velocity,
  remodeling and arrhythmogenesis. Cells  \textbf{10}(11), ~2923 (2021)

\bibitem{journal/Lancet/kaptoge2019}
Kaptoge, S., Pennells, L., De~Bacquer, D., Cooney, M.T., Kavousi, M., Stevens,
  G., Riley, L.M., Savin, S., Khan, T., Altay, S., et~al.: {World Health
  Organization} cardiovascular disease risk charts: revised models to estimate
  risk in 21 global regions. The Lancet Global Health  \textbf{7}(10),
  e1332--e1345 (2019)

\bibitem{journal/IF/martinez2021}
Martinez-Navarro, H., Zhou, X., Bueno-Orovio, A., Rodriguez, B.:
  Electrophysiological and anatomical factors determine arrhythmic risk in
  acute myocardial ischaemia and its modulation by sodium current availability.
  Interface Focus  \textbf{11}(1),  20190124 (2021)

\bibitem{conf/STACOM/meister2020}
Meister, F., Passerini, T., Audigier, C., Lluch, {\`E}., Mihalef, V., Ashikaga,
  H., Maier, A., Halperin, H., Mansi, T.: Graph convolutional regression of
  cardiac depolarization from sparse endocardial maps. In: International
  Workshop on Statistical Atlases and Computational Models of the Heart. pp.
  23--34. Springer (2020)

\bibitem{journal/FiP/minchole2019}
Minchol{\'e}, A., Zacur, E., Ariga, R., Grau, V., Rodriguez, B.: {MRI}-based
  computational torso/biventricular multiscale models to investigate the impact
  of anatomical variability on the {ECG QRS} complex. Frontiers in Physiology
  p.~1103 (2019)

\bibitem{journal/NRC/niederer2019}
Niederer, S.A., Lumens, J., Trayanova, N.A.: Computational models in
  cardiology. Nature Reviews Cardiology  \textbf{16}(2),  100--111 (2019)

\bibitem{journal/PCE/park2012}
Park, K.M., Kim, Y.H., Marchlinski, F.E.: Using the surface electrocardiogram
  to localize the origin of idiopathic ventricular tachycardia. Pacing and
  Clinical Electrophysiology  \textbf{35}(12),  1516--1527 (2012)

\bibitem{journal/TBME/pashaei2011}
Pashaei, A., Romero, D., Sebastian, R., Camara, O., Frangi, A.F.: Fast
  multiscale modeling of cardiac electrophysiology including purkinje system.
  IEEE transactions on biomedical engineering  \textbf{58}(10),  2956--2960
  (2011)

\bibitem{journal/EP/pezzuto2021}
Pezzuto, S., Prinzen, F.W., Potse, M., Maffessanti, F., Regoli, F., Caputo,
  M.L., Conte, G., Krause, R., Auricchio, A.: Reconstruction of
  three-dimensional biventricular activation based on the 12-lead
  electrocardiogram via patient-specific modelling. EP Europace
  \textbf{23}(4),  640--647 (2021)

\bibitem{journal/ACM/si2015}
Si, H.: {TetGen}, a {Delaunay}-based quality tetrahedral mesh generator. ACM
  Transactions on Mathematical Software (TOMS)  \textbf{41}(2),  1--36 (2015)

\bibitem{journal/PCE/taneja2001}
Taneja, T., Windhagen~Mahnert, B., Passman, R., Goldberger, J., Kadish, A.:
  Effects of sex and age on electrocardiographic and cardiac
  electrophysiological properties in adults. Pacing and Clinical
  Electrophysiology  \textbf{24}(1),  16--21 (2001)

\bibitem{conf/3DV/yuan2018}
Yuan, W., Khot, T., Held, D., Mertz, C., Hebert, M.: {PCN}: Point completion
  network. In: 2018 International Conference on 3D Vision. pp. 728--737. IEEE
  (2018)

\end{thebibliography}

\end{document}